\documentclass[conference]{IEEEtran}
\IEEEoverridecommandlockouts

\usepackage{cite}
\usepackage{amsmath,amssymb}
\usepackage{graphicx}
\usepackage{booktabs}
\usepackage{tabularx}
\usepackage{listings}
\usepackage{url}
\usepackage{placeins}
\usepackage{flafter}

\newcolumntype{Y}{>{\raggedright\arraybackslash}X}

\lstdefinestyle{sql}{
	language=SQL,
	basicstyle=\fontsize{7.1pt}{8.6pt}\selectfont\ttfamily,
	keywordstyle=\bfseries,
	columns=fullflexible,
	keepspaces=true,
	breaklines=true,
	showstringspaces=false,
	aboveskip=3pt,
	belowskip=2pt,
	xleftmargin=1pt,
}

\newcommand{\HDFSRSAUC}{0.999}
\newcommand{\HDFSRSSD}{$<$.001}
\newcommand{\HDFSAEAUC}{1.000}
\newcommand{\HDFSAESD}{$<$.001}
\newcommand{\HDFSFA}{0.041}
\newcommand{\HDFSDR}{1.000}
\newcommand{\HDFSFAOne}{0.010}
\newcommand{\HDFSDROne}{1.000}
\newcommand{\HDFSFATen}{0.098}
\newcommand{\HDFSDRTen}{1.000}
\newcommand{\BGLRSAUC}{0.923}

\newcommand{\BGLFA}{0.038}
\newcommand{\BGLDR}{0.760}
\newcommand{\BGLFAOne}{0.006}
\newcommand{\BGLDROne}{0.490}
\newcommand{\BGLFATen}{0.080}
\newcommand{\BGLDRTen}{0.767}
\newcommand{\BGLChronRS}{0.699}
\newcommand{\BGLChronAE}{0.864}
\newcommand{\BGLChronAESD}{0.001}

\newcommand{\ThunderbirdFA}{0.046}
\newcommand{\ThunderbirdDR}{0.213}
\newcommand{\ThunderbirdFAOne}{0.008}
\newcommand{\ThunderbirdDROne}{0.024}
\newcommand{\ThunderbirdFATen}{0.094}
\newcommand{\ThunderbirdDRTen}{0.527}
\newcommand{\ThunderbirdChronRS}{0.983}
\newcommand{\ThunderbirdChronAE}{0.949}
\newcommand{\ThunderbirdChronAESD}{0.001}
\newcommand{\FixedRareRS}{0.787}
\newcommand{\FixedRareAE}{0.771}

\newcommand{\FixedHeadRS}{0.398}
\newcommand{\FixedHeadAE}{0.594}
\newcommand{\FixedTransferPct}{18.5}
\newcommand{\FixedMaxGapThroughSeventyFive}{0.016}

\newcommand{\CTURSRecall}{0.228}

\newcommand{\CTUTrafficRecall}{0.644}

\newcommand{\CostRSOnline}{19.5}
\newcommand{\CostRSOnlineP}{60.4}
\newcommand{\CostAEOnline}{43.5}
\newcommand{\CostAEOnlineP}{65.7}
\newcommand{\CostRSSetup}{0.009}
\newcommand{\CostRSSetupP}{0.011}
\newcommand{\CostAESetup}{2.279}
\newcommand{\CostAESetupP}{20.104}
\newcommand{\CostRSState}{5,160}
\newcommand{\CostAEState}{706,264}
\newcommand{\CostStateRatio}{137}
\newcommand{\DuckDBSixtyFour}{12.2}

\newcommand{\DDBZeroSix}{0.428}
\newcommand{\DDBZeroSixP}{0.448}
\newcommand{\DDBZeroSixMem}{0.12}
\newcommand{\DDBThree}{0.813}
\newcommand{\DDBThreeP}{0.855}
\newcommand{\DDBThreeMem}{0.42}
\newcommand{\DDBSixteen}{2.972}
\newcommand{\DDBSixteenP}{3.288}
\newcommand{\DDBSixteenMem}{1.27}
\newcommand{\DDBThirtyTwo}{5.820}
\newcommand{\DDBThirtyTwoP}{5.993}
\newcommand{\DDBThirtyTwoMem}{2.27}
\newcommand{\DDBSixtyFourP}{12.881}
\newcommand{\DDBSixtyFourMem}{5.09}
\newcommand{\ADXZeroSix}{0.900}

\newcommand{\ADXSixtyFour}{11.000}

\newcommand{\ADXSixtyFourMem}{9.00}

\def\BibTeX{{\rm B\kern-.05em{\sc i\kern-.025em b}\kern-.08em
T\kern-.1667em\lower.7ex\hbox{E}\kern-.125emX}}

\begin{document}

\title{RankShift: In-Database Detection and Explanation of Categorical Shifts}

\author{\IEEEauthorblockN{Omair Shafi Ahmed}
\IEEEauthorblockA{\textit{Microsoft} \\
oshafiahmed@microsoft.com}}

\maketitle

\begin{abstract}
A login service can receive its usual number of failed sign-ins while one source
grows from 2\% to 30\% of them. The same pattern appears in system logs when a
rare event template becomes common while the message rate stays stable. These
events change which categories are active without changing how many events
occur. RankShift detects such changes inside the analytical database that
stores the data. It compares each window's category shares with a benign
reference using a Pearson score whose terms identify the categories responsible
for the change. The same query returns the score, calibrated alert, and largest
increasing contributions.

We evaluate RankShift on HDFS, BGL, and Thunderbird. It matches the
count-vector autoencoder within 0.001 AUROC on HDFS
(\HDFSRSAUC\ versus \HDFSAEAUC) and leads on Thunderbird
(\ThunderbirdChronRS\ versus \ThunderbirdChronAE). In a controlled
fixed-volume experiment, RankShift detects rare-category shifts that are
invisible to event-count monitoring, reaching \FixedRareRS\ AUROC compared
with \FixedRareAE\ for the autoencoder. Across all three corpora, observed
false-alarm rates track the requested operating levels. RankShift requires no
model training or inference service, and the autoencoder's deployed state is
\CostStateRatio$\times$ larger.
\end{abstract}

\begin{IEEEkeywords}
categorical data, anomaly detection, distribution shift, data streams,
in-database analytics
\end{IEEEkeywords}

\section{Introduction}

\begin{figure*}[t]
\centering
\includegraphics[width=0.94\textwidth]{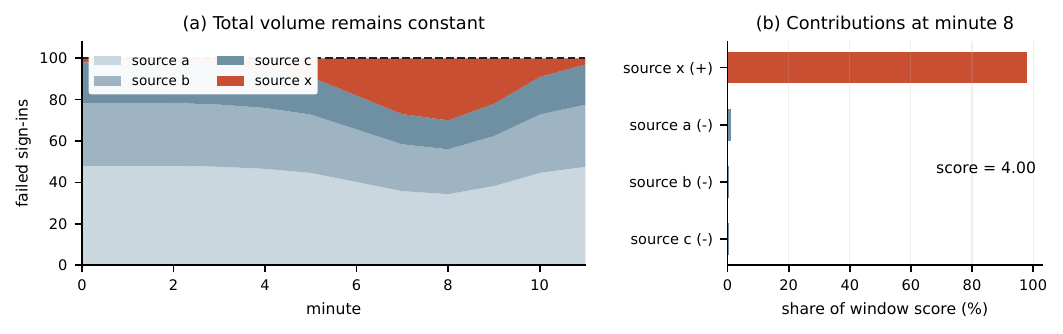}
\caption{Illustrative fixed-volume shift in failed sign-ins. (a)~Each minute
contains 100 failed sign-ins, while source~x rises from 2\% to 30\% of the
total. A monitor of total volume remains flat. (b)~At minute~8, source~x
accounts for 98\% of the RankShift score (3.92 of 4.00). The signs show whether
each source gained or lost share.}
\label{fig:story}
\end{figure*}

Monitoring systems often summarize each time window by request rate, error
count, or latency. These measures can remain stable while the shares of source
addresses, routes, event templates, or error codes change sharply.
Figure~\ref{fig:story} illustrates this pattern in a failed-sign-in stream.
Changes in categorical composition can reveal an emerging source, a failing
component, or a new workload pattern that aggregate monitoring misses. A useful
monitor must detect the change and identify which categories gained or lost
share.

Existing log detectors require a preprocessing, fitting, and scoring pipeline.
Count-vector methods first group parsed events into windows and fit a model over
the resulting vectors \cite{xu2009detecting}. The autoencoder evaluated here
learns normalization statistics from benign fitting windows, trains an
encoder-decoder, and scores later windows by reconstruction error. Sequence
detectors construct ordered event-template sequences and train next-event or
masked-event predictors \cite{du2017deeplog,guo2021logbert}. Deployment retains
preprocessing statistics, learned weights, and an inference runtime.

Distribution-change methods use different operational state. QuantTree
estimates a partition from reference observations, while MStream maintains hash
sketches and decayed counts \cite{boracchi2018quanttree,bhatia2021mstream}.
They run as dedicated streaming algorithms outside the analytical event table.

This paper makes three contributions:

\begin{itemize}
\item \textbf{Monitoring method.} RankShift uses category-level Pearson
contributions to score a window and identify the categories driving the change.
Benign score ranks calibrate the alert.
\item \textbf{Evaluation.} We evaluate RankShift on three real log datasets and
use fixed-total shifts to show when it leads or trails a count-vector
autoencoder.
\item \textbf{Native execution.} It implements equivalent NumPy, DuckDB SQL,
and ADX KQL semantics and measures correctness, event-volume scaling, setup,
online cost, and deployed state.
\end{itemize}

\section{Related Work}

\textbf{Distribution monitoring.} Pearson's chi-square goodness-of-fit test
compares observed and theoretical frequencies across groups
\cite{pearson1900criterion}. Its statistic sums one normalized squared deviation
for each group, allowing the total discrepancy to be traced to individual
groups. Lakhina et al. analyzed distributions of source addresses, destination
addresses, and ports in network flows \cite{lakhina2005mining}. They summarized
these distributions with entropy, detected anomalies beyond those found by
volume-based methods, and clustered the results for unsupervised anomaly
classification. QuantTree learns a multivariate histogram with prescribed bin
probabilities from stationary reference data, then tests later batches using
the number of observations entering each bin \cite{boracchi2018quanttree}.
RankShift retains operational category names, such as source addresses and
event templates, and compares their shares so a uniform change in total count
leaves the score unchanged.

\textbf{Log and stream detectors.} Log detectors differ in the relationships
they learn. Xu et al. represent each window as event-template counts and use PCA
to learn correlations among event types \cite{xu2009detecting}. DeepLog learns
which event template should follow a preceding sequence, while LogBERT learns
normal sequence context through masked-event prediction and a one-class
objective \cite{du2017deeplog,guo2021logbert}. Stream detectors operate on other
structures. MStream captures correlations across the categorical and numerical
attributes of each record, whereas MIDAS detects sudden bursts of repeated
source-destination edges \cite{bhatia2021mstream,bhatia2020midas}. RankShift
addresses changes in the relative frequencies within one named categorical
field. Event order, cross-field relationships, and graph structure lie outside
this scope.

\textbf{Explaining anomalies.} Existing methods usually explain a result
produced by another detector. SHAP attributes a model prediction to its input
features, with computation and exactness determined by the model and chosen
explainer \cite{lundberg2017unified}. ACE and DeepAID similarly explain anomaly
scores after a detector has produced them
\cite{zhang2019ace,han2021deepaid}. MacroBase instead identifies attribute
values that are enriched among records already classified as outliers, while
weighted log odds ranks terms that distinguish two corpora
\cite{bailis2017macrobase,monroe2008fightin}. RankShift requires no separate
explainer. Its category contributions are computed as part of the score, sum
exactly to that score, and retain operational values such as source addresses
or event templates.

\textbf{In-database analytics.} MADlib brings statistical analysis and machine
learning into relational databases, allowing model fitting and inference to run
where the data is stored \cite{hellerstein2012madlib}. DuckDB provides an
embedded analytical database that executes SQL directly within an application
process \cite{raasveldt2019duckdb}. These systems establish the value of
performing analytics without moving data into a separate processing service.
RankShift applies this principle to categorical monitoring and returns both an
alert and its category-level decomposition from relational operations.

\textbf{Alert calibration.} An anomaly score does not by itself define when to
raise an alert. Split-conformal methods compare a new score with scores from
held-out calibration examples and convert its rank into a p-value
\cite{vovk2005algorithmic,angelopoulos2023conformal,bates2023testing}. When
benign calibration and future windows are exchangeable, alerting at level
$\alpha$ controls the marginal false-alarm probability at $\alpha$. RankShift
uses benign calibration windows to set this operating level without assuming a
parametric distribution for its scores.

\section{RankShift}\label{sec:method}

\subsection{Categories and reference shares}

RankShift monitors one categorical field in timestamped events grouped into
fixed time windows. Before monitoring begins, every field value is mapped to one
of $C$ categories. A field with a known vocabulary, such as an HTTP status code,
can use its values directly. For unstructured logs, a parser and template
dictionary can be learned from historical data and then frozen; later messages
that match no template map to \texttt{OTHER}. A fixed-size hash mapping is
another option for large vocabularies. The mapping determines how precisely
RankShift can identify a driver: values merged into \texttt{OTHER} or the same
hash bucket receive one combined contribution.

Benign history establishes the expected category mix. Let $r_i$ be the number
of reference events assigned to category $i$, and let $R=\sum_i r_i$ be the
total number of reference events. A category absent from this history has a raw
share of zero, which cannot be used in the Pearson score because the score
divides by the reference share. RankShift therefore spreads a small pseudocount
across all $C$ categories:
\begin{equation}
q_i=\frac{r_i+\tau/C}{R+\tau},\qquad \tau>0.
\label{eq:reference}
\end{equation}
The resulting reference shares $q_i$ are positive and sum to one. Assigning
$\tau/C$ to each category keeps the total smoothing equal to $\tau$ regardless
of vocabulary size. We use $\tau=1$ throughout, adding one count of total mass
divided equally across the category set.

\subsection{Window score and category contributions}

For a window containing $N>0$ events, let $s_i$ be the number assigned to
category $i$. Its observed category share is $p_i=s_i/N$, where
$N=\sum_i s_i$. RankShift compares the observed shares $p$ with the reference
shares $q$:
\begin{equation}
c_i=\frac{(p_i-q_i)^2}{q_i},\qquad
X=\sum_{i=1}^{C}c_i .
\label{eq:score}
\end{equation}
The total $X$ is the Pearson divergence between the observed and reference
compositions. Each term $c_i$ measures how much category $i$ contributes to
that difference. For the same change in share, a category that was rare in the
reference receives a larger contribution because $q_i$ appears in the
denominator. Every contribution is nonnegative, and the complete set sums
exactly to $X$. Squaring removes direction, so RankShift also records the sign
of $p_i-q_i$: positive values gained share and negative values lost share.

For triage, RankShift reports the $k$ largest contributions among categories
whose shares increased. Categories that lost share and contributions below rank
$k$ remain available in the full table but do not appear in this ranked view.

RankShift normalizes each window by its total event count before scoring. If
every category count grows or shrinks by the same factor, the observed shares
$p_i$ remain unchanged, and so do the category contributions and total score
$X$. The score therefore isolates changes in category mix from uniform changes
in traffic volume. It can detect a redistribution while the total event rate
remains stable, as in Figure~\ref{fig:story}. A surge or drop that preserves the
same category shares produces no RankShift signal and requires a separate
volume monitor.

\subsection{Alert calibration}

Once the reference shares are fixed, a separate set of benign windows defines
the normal range of scores. Let $T_1,\ldots,T_n$ be the scores of $n\geq1$
calibration windows. For a new window with score $X$, RankShift computes
\begin{equation}
\widehat p=\frac{1+|\{j:T_j\ge X\}|}{n+1}.
\label{eq:conformal}
\end{equation}
The numerator counts calibration windows whose scores are at least as large as
$X$. The additional one includes the new score in its finite-sample rank and
prevents a p-value of zero. A small $\widehat p$ means that few benign
calibration windows produced a score as large as the new window.

RankShift raises an alert when $\widehat p\leq\alpha$. If the calibration scores
and future benign scores are exchangeable after the reference is fixed, the
marginal false-alarm probability is at most $\alpha$
\cite{vovk2005algorithmic,bates2023testing}. The $\geq$ comparison counts tied
calibration scores against the alert. A deployment with a fixed $\alpha$ can
store the corresponding score cutoff instead of the complete calibration
sample.

\section{Database Execution}\label{sec:execution}

After the category mapping is fixed, RankShift's deployed state consists of $C$
reference shares and either $n$ sorted calibration scores or a fixed alert
cutoff and comparison rule. Its scoring and calibration procedure has no model
training phase, learned weights, model-inference runtime, or GPU requirement.
While a window is open, each event increments one category count. For a window
with $a$ active categories, the sparse score requires $O(a)$ work, and selecting
the largest $k$ increasing contributions requires $O(a\log k)$ work. A sorted
calibration sample supports p-value lookup in $O(\log n)$ time; a fixed alert
cutoff reduces the decision to $O(1)$.

RankShift begins by counting events by cohort, window, and category. The
database materializes this grouped table once, allowing reference construction,
window scoring, calibration, and driver ranking to reuse the same counts without
rescanning the raw events. A dense implementation would create $C$ rows for
every window, including categories with zero events. RankShift instead uses the
identity
\[
X=\sum_{i:p_i>0}\frac{p_i^2}{q_i}-1,
\]
which evaluates the exact score using only categories present in the window.
Appendix~\ref{app:identities} gives the derivation. Reference counts produce the
smoothed shares $q_i$. When a present category has no reference count, the query
assigns it the smoothed floor $(\tau/C)/(R+\tau)$ from
Eq.~\eqref{eq:reference}.

After scoring each test window, the query counts calibration scores at least as
large as the test score to produce the conformal p-value and alert. It then
keeps categories whose shares increased, orders them by contribution and
category identifier, and returns the largest $k$. Each result contains the
window score, p-value, alert, selected category values, and their contributions.
Dividing the sum of the returned contributions by $X$ gives the fraction of the
score represented by the ranked view. Appendix~\ref{app:query} prints the
complete DuckDB SQL executed in the experiments.

Conformal alerts depend on the ordering of window scores. Small floating-point
differences between query engines can change a tie or move a score across the
alert boundary. RankShift therefore separates the reported score from the value
used for decisions. It returns the unrounded score $X$, but rounds a nonnegative
copy to six decimal places, with exact halves rounded upward, before comparing
calibration and test scores. DuckDB and ADX use the same rule.

We compared both query engines with the NumPy reference implementation on a
30-window BGL fixture. The maximum absolute score difference was
$3\times10^{-8}$ for DuckDB and $8\times10^{-9}$ for ADX. Both engines produced
identical p-values, alerts, contribution values, and category ordering.
Additional tests at exact rounding boundaries confirmed the same decision rule.

\section{Evaluation}\label{sec:evaluation}

The experiments test different parts of the RankShift result. HDFS, BGL, and
Thunderbird measure anomaly detection and false-alarm calibration on natural log
windows. A fixed-total experiment changes category shares while preserving every
window's event count, isolating composition from volume. DuckDB and ADX measure
native query execution. CTU-13 tests whether a bounded category mapping preserves
enough information to identify raw attack sources. In every comparison, the
category mapping is fixed before the test period, and RankShift and its baseline
receive the same count matrix and test windows. Table~\ref{tab:contract}
summarizes these protocols.

\begin{table*}[t]
\centering
\caption{Datasets and evaluation protocols. Category vocabularies are fixed
before testing. Alert evaluation uses disjoint benign reference, calibration,
and test cohorts.}
\label{tab:contract}
\footnotesize
\setlength{\tabcolsep}{3.5pt}
\renewcommand{\arraystretch}{1.08}
\begin{tabularx}{\textwidth}{@{}l >{\raggedright\arraybackslash}p{0.10\textwidth} Y Y Y >{\raggedright\arraybackslash}p{0.13\textwidth}@{}}
\toprule
dataset & window / categories & vocabulary construction & detection split & alert split & reported result \\
\midrule
HDFS & block / 29 & fixed event identifiers & benign half $\rightarrow$ held-out benign + anomalies & random benign reference/calibration/test split & detection, alert \\
BGL & hour / 644 & first-half Drain, frozen + OTHER & first-half benign $\rightarrow$ complete second half & second-half benign reference/calibration/test split & detection, alert, redistribution \\
Thunderbird & minute / 1{,}025 & first-half benign top-1{,}024 + OTHER & first-half benign $\rightarrow$ complete second half & second-half benign reference/calibration/test split & detection, alert \\
CTU-13 & minute / 2{,}049 & first-half benign top-2{,}048 + OTHER & first-half benign $\rightarrow$ second-half attacks & -- & raw-source recovery \\
\bottomrule
\end{tabularx}
\end{table*}

\subsection{Data and protocol}

HDFS, BGL, and Thunderbird are public system-log datasets distributed through
Loghub~\cite{zhu2023loghub}. HDFS records operations on Hadoop file-system
blocks. Events associated with the same block form one sample, producing
575,061 block samples over 29 event identifiers; the published block label
defines whether each sample is anomalous. BGL and Thunderbird contain
timestamped supercomputer logs with message-level anomaly labels. We group BGL
into 3,619 one-hour windows and the first ten million Thunderbird messages into
30,120 one-minute windows. A time window is anomalous when it contains at least
one labeled anomalous message.

Category vocabularies use only information available before testing. For BGL,
Drain learns templates from the first half of the log and is then frozen
\cite{he2017drain}. Its 643 learned templates plus \texttt{OTHER} produce 644
categories. For Thunderbird, the vocabulary retains the 1,024 most frequent
benign templates from the first half and maps every remaining template to
\texttt{OTHER}. Both frozen mappings are applied unchanged throughout the
second-half tests.

Detection comparisons use the same benign fitting data and test samples for
both methods. HDFS uses five paired random splits. In each split, one benign
half defines the RankShift reference and trains the autoencoder; the remaining
benign samples and all anomalous samples form the shared test set. BGL and
Thunderbird preserve time order: first-half benign windows define the reference
and train the autoencoder, and every second-half window is tested. Autoencoder
training is repeated with five random seeds.

The autoencoder receives the same category-count vectors as RankShift. Its
preprocessing applies $\log(1+x)$ and estimates each feature's mean and standard
deviation from the benign fitting windows only. The encoder contains at most 128
hidden units and a 32-unit bottleneck, followed by a symmetric decoder. Training
minimizes reconstruction error on benign windows, and mean reconstruction error
scores each test window.

Alert calibration is evaluated separately from detection ranking. HDFS benign
samples are divided into disjoint reference, calibration, and test cohorts. For
BGL and Thunderbird, the category mapping is learned from the first half, then
the later benign windows are divided into the three cohorts. This measures false
alarms under the later operating distribution without allowing any test label
to influence the reference or calibration scores.

\subsection{Detection and alert calibration on labeled logs}

Table~\ref{tab:detection} compares RankShift and the autoencoder on the same
test samples.
RankShift and the autoencoder both separate HDFS anomalies almost perfectly,
reaching \HDFSRSAUC\ and \HDFSAEAUC\ AUROC. On the complete second half of BGL,
RankShift reaches \BGLChronRS\ compared with \BGLChronAE\ for the autoencoder.
On the complete second half of Thunderbird, RankShift reaches
\ThunderbirdChronRS\ compared with \ThunderbirdChronAE. The opposing BGL and
Thunderbird results motivate the fixed-total experiment below, which tests how
performance changes when shifted activity enters rare or common categories.

\begin{table}[!ht]
\centering
\caption{Detection AUROC on shared test samples. HDFS reports mean (standard
deviation) over five paired splits. BGL and Thunderbird use first-half benign
windows for setup and the complete second half for testing; autoencoder values
report mean (standard deviation) over five training seeds.}
\label{tab:detection}
\setlength{\tabcolsep}{3.5pt}
\begin{tabular}{@{}llcc@{}}
\toprule
dataset & test period & RankShift & autoencoder \\
\midrule
HDFS & held-out & \HDFSRSAUC\ (\HDFSRSSD) & \HDFSAEAUC\ (\HDFSAESD) \\
BGL & second half & \BGLChronRS & \BGLChronAE\ (\BGLChronAESD) \\
Thunderbird & second half & \ThunderbirdChronRS & \ThunderbirdChronAE\ (\ThunderbirdChronAESD) \\
\bottomrule
\end{tabular}
\end{table}

Table~\ref{tab:calibration} reports mean alert rates over five random
partitions. Within each partition, the benign reference, calibration, and test
cohorts are disjoint. At the requested 0.05 false-alarm level, the observed
rates are \HDFSFA\ on HDFS, \BGLFA\ on BGL, and \ThunderbirdFA\ on Thunderbird.
HDFS detects every anomaly at all three alert levels. BGL detection rises from
\BGLDROne\ at $\alpha=0.01$ to \BGLDR\ at $\alpha=0.05$, then changes little at
$\alpha=0.10$. Thunderbird detection rises from \ThunderbirdDROne\ to
\ThunderbirdDR\ and \ThunderbirdDRTen\ across the same levels.

Thunderbird's \ThunderbirdChronRS\ AUROC and its \ThunderbirdDR\ detection rate
at $\alpha=0.05$ measure different properties. AUROC evaluates score ordering
across all possible thresholds. The detection rate measures recall at one
threshold chosen to limit benign false alarms. Strong overall ordering can
therefore coexist with low recall at a strict operating point.

\begin{table}[!ht]
\centering
\caption{Calibrated alert rates averaged over five random partitions. Within
each partition, the benign reference, calibration, and test cohorts are
disjoint. False-alarm rate is measured on benign test windows; detection rate
is measured on anomalous windows.}
\label{tab:calibration}
\setlength{\tabcolsep}{7pt}
\begin{tabular}{@{}lccc@{}}
\toprule
dataset & $\alpha$ & false alarm & detection \\
\midrule
HDFS & .01 & \HDFSFAOne & \HDFSDROne \\
	& .05 & \HDFSFA & \HDFSDR \\
	& .10 & \HDFSFATen & \HDFSDRTen \\
\addlinespace[2pt]
BGL  & .01 & \BGLFAOne & \BGLDROne \\
	& .05 & \BGLFA & \BGLDR \\
	& .10 & \BGLFATen & \BGLDRTen \\
\addlinespace[2pt]
Thunderbird & .01 & \ThunderbirdFAOne & \ThunderbirdDROne \\
		  & .05 & \ThunderbirdFA & \ThunderbirdDR \\
		  & .10 & \ThunderbirdFATen & \ThunderbirdDRTen \\
\bottomrule
\end{tabular}
\end{table}
\FloatBarrier

RankShift trails the autoencoder on BGL and leads on Thunderbird. The next
subsection tests whether this contrast is related to how often the changing
categories appeared in benign history.

\subsection{Controlled synthetic redistribution}

For each of five seeds, we randomly divide benign BGL windows from the second
half of the log into two equal groups after the parser has been frozen. The
setup group defines RankShift's reference shares and trains the autoencoder.
From the evaluation group, we select 400 windows containing at least 20 events.

Each selected window remains unchanged and serves as a benign example. We
create a synthetic anomaly by copying the window and moving some of its event
counts between categories. Categories observed in the setup group are ranked by
their total counts. The least frequent 10\% form the rare-recipient pool, and
the most frequent 10\% form the common-recipient pool. For each window, we
sample five recipient categories from each pool.

We attempt to move 30\% of the window's events. Counts are removed from its
largest active categories outside the selected recipients and reassigned to the
recipients. The fraction sent to common recipients varies from 0\% to 100\% in
25-point steps; the remainder goes to rare recipients. Limited donor counts and
integer rounding produce an average realized transfer of
\FixedTransferPct\%. No events are added or removed, so every synthetic anomaly
has exactly the same total as its original benign window. A volume-only
detector therefore assigns identical scores to both classes and has 0.500
AUROC.

\begin{figure}[t]
\centering
\includegraphics[width=\columnwidth]{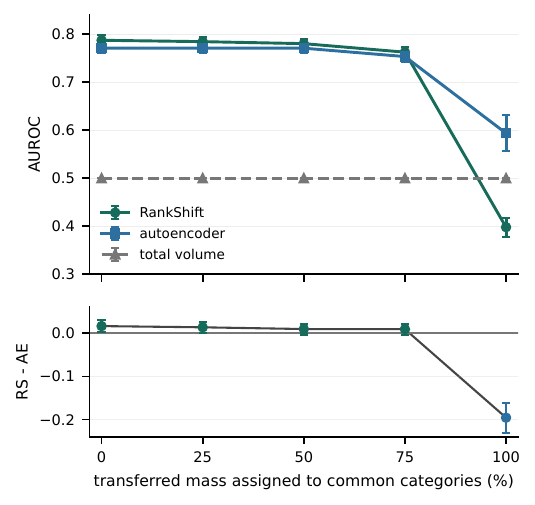}
\caption{Detection under controlled fixed-total redistribution on BGL. The
horizontal axis gives the fraction of transferred events assigned to common
recipient categories; the remainder enters rare recipient categories. The
upper panel reports mean AUROC over five paired seeds. The lower panel reports
RankShift minus autoencoder AUROC, with positive values favoring RankShift.
Error bars show one standard deviation. Every synthetic anomaly has the same
event count as its original benign window, giving the volume baseline 0.500
AUROC.}
\label{fig:fixed-total}
\end{figure}

At 0\% common allocation, when all transferred events enter rare recipients,
RankShift reaches \FixedRareRS\ AUROC compared with \FixedRareAE\ for the
autoencoder. The methods remain within \FixedMaxGapThroughSeventyFive\ AUROC
through 75\% common allocation. At 100\%, RankShift falls to \FixedHeadRS\ while
the autoencoder reaches \FixedHeadAE{}. RankShift emphasizes changes in rare
categories because each contribution is divided by the category's benign
reference share $q_i$. The same absolute change therefore produces a larger
contribution for a rare category than for a common one. The \FixedHeadRS\ result
means that synthetic anomalies tend to receive lower RankShift scores than the
original benign windows when all transferred events enter common recipients.
Under this controlled intervention, RankShift responds most strongly when
activity moves into categories that were rare in benign history and loses
sensitivity as the redistribution becomes confined to common categories.

\subsection{Execution and cost}

The execution benchmark measures the complete RankShift query, including the
score, conformal p-value, alert, and five largest increasing contributions.
Event rows are generated and loaded before timing begins. The benchmark varies
the number of windows while holding the workload structure fixed: 5,000
supported categories, 80 active categories per window, and four events per
active category. Each reported value is the median of five warm query
executions.

On the measured DuckDB virtual machine, median query time rises from
\DDBZeroSix\ seconds for 0.64 million events to \DuckDBSixtyFour\ seconds for
64 million events. The largest run processes approximately 5.2 million events
per second. Every query profile contains one scan of the event table.

ADX median query time rises from \ADXZeroSix\ seconds for 0.64 million events
to \ADXSixtyFour\ seconds for 64 million events. The largest run processes
approximately 5.8 million events per second. DuckDB and ADX run on different
infrastructure. Their measurements characterize scaling within each engine and
do not support a direct speed ranking between them.

\begin{figure}[t]
\centering
\includegraphics[width=\columnwidth]{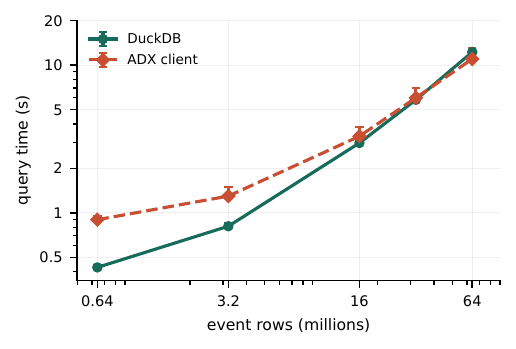}
\caption{Complete RankShift query time as event volume increases. Lines report
median client wall time over five warm runs; upper bars report p95. DuckDB
and ADX complete workloads through 64 million events.}
\label{fig:execution}
\end{figure}

\begin{table*}[t]
\centering
\caption{Native-query scaling under the same logical workload. Client entries
are median (p95) seconds over five warm runs. Peak memory is the maximum
observed across those runs; ADX memory is per node.}
\label{tab:engines}
\setlength{\tabcolsep}{7pt}
\begin{tabular}{@{}rcccc@{}}
\toprule
events (M) & DuckDB client & DuckDB GiB & ADX client & ADX GiB/node \\
\midrule
0.64 & \DDBZeroSix\ (\DDBZeroSixP) & \DDBZeroSixMem & 0.900 (0.950) & 0.12 \\
3.2  & \DDBThree\ (\DDBThreeP) & \DDBThreeMem & 1.300 (1.500) & 0.50 \\
16   & \DDBSixteen\ (\DDBSixteenP) & \DDBSixteenMem & 3.300 (3.800) & 2.50 \\
32   & \DDBThirtyTwo\ (\DDBThirtyTwoP) & \DDBThirtyTwoMem & 6.000 (7.000) & 5.00 \\
64   & \DuckDBSixtyFour\ (\DDBSixtyFourP) & \DDBSixtyFourMem & 11.000 (13.000) & 9.00 \\
\bottomrule
\end{tabular}
\end{table*}

At 64 million events, DuckDB uses \DDBSixtyFourMem\ GiB of peak buffer memory
and ADX uses \ADXSixtyFourMem\ GiB of peak query memory per node.

A separate benchmark isolates method cost from database infrastructure. It runs
RankShift and the autoencoder in the same process on the 644-category BGL
representation. Both methods return a score, conformal p-value, alert, and five
ranked categories. Values are medians over five complete runs; each online
measurement contains two warmups and five timed repetitions.

\begin{table}[!ht]
\centering
\caption{Matched same-process cost on reference-frozen BGL. Times are median
(p95) over five complete runs.}
\label{tab:cost}
\setlength{\tabcolsep}{5pt}
\begin{tabular}{@{}lcc@{}}
\toprule
measure & RankShift & autoencoder \\
\midrule
online ($\mu$s/window) & \CostRSOnline\ (\CostRSOnlineP) & \CostAEOnline\ (\CostAEOnlineP) \\
setup (s) & \CostRSSetup\ (\CostRSSetupP) & \CostAESetup\ (\CostAESetupP) \\
deployed state (bytes) & \CostRSState & \CostAEState \\
\bottomrule
\end{tabular}
\end{table}

RankShift scores a window in a median of \CostRSOnline\ microseconds, compared
with \CostAEOnline\ microseconds for the autoencoder. Its median setup time is
\CostRSSetup\ seconds, compared with \CostAESetup\ seconds. RankShift retains
\CostRSState\ bytes of deployed state, while the autoencoder retains
\CostAEState\ bytes, making RankShift's state \CostStateRatio$\times$ smaller.
Online and setup times vary across complete runs; the retained-state sizes are
fixed by the measured implementations.

\section{Operational Scope}

RankShift is designed to detect changes in category composition. A proportional
increase or decrease across all categories leaves the score unchanged because
the category shares remain unchanged. A deployment that must also detect
changes in total traffic should pair RankShift with a volume monitor.

Window size determines how much category shares vary under normal conditions.
For example, a 30-to-70 split observed across 10 events is less stable than the
same split observed across 10,000 events. Reference, calibration, and monitored
windows should therefore use the same duration and represent similar
event-volume ranges. The calibration scores then reflect the normal sampling
variation of the deployed windows.

The benign category distribution can also change over time. On BGL, RankShift
reaches \BGLRSAUC\ AUROC under a random held-out split and \BGLChronRS\ when the
first half defines the reference and the complete second half is tested. This
decline shows the effect of evaluating against an older reference. Reference
shares and calibration scores should be refreshed when sustained benign changes
alter the operating distribution. The conformal guarantee requires future
benign scores to be exchangeable with the calibration scores.

The category vocabulary determines the precision of the reported drivers. In
eight CTU-13 captures, every labeled attack source is absent from the
2,048-address reference vocabulary and maps to \texttt{OTHER}
\cite{garcia2014ctu13}. Across 1,909 later attack windows, RankShift recovers
\CTURSRecall\ of true sources among its first five candidates, compared with
\CTUTrafficRecall\ for current-count ranking. RankShift computes an exact
contribution for \texttt{OTHER} as a group. It cannot identify an individual
address within that group. Future versions can address this limitation with a
hierarchical vocabulary that refines an anomalous group into progressively
narrower subgroups and raw values while retaining bounded top-level state.

\newpage
\section{Conclusion}

RankShift turns a classical Pearson comparison into an operational monitor for
categorical streams. Its per-category contributions produce the anomaly score
and identify which values gained or lost share. Benign calibration converts the
score into an alert. The complete calculation runs inside the analytical
database without a trained detector or separate explanation system.

The evaluation defines RankShift's operating range. It responds strongly when
activity moves into categories that were rare in benign history and loses
sensitivity when redistribution remains among common categories. Its reported
drivers are exact at the resolution of the deployed vocabulary; values merged
into \texttt{OTHER} require a finer or hierarchical representation. RankShift
targets internet-scale categorical telemetry where model training, inference,
and movement of raw events to a separate detector are infeasible. It executes
inside the analytical database's optimized counting path, providing low-latency
monitoring together with category-level drivers. Changes in total activity
remain a separate monitoring channel.

\appendices
\section{Supporting Identities}\label{app:identities}

The database calculates the exact score without creating rows for categories
absent from a window. Expanding Eq.~\eqref{eq:score} gives
\begin{equation}
\begin{aligned}
X
&=\sum_i\frac{(p_i-q_i)^2}{q_i} \\
&=\sum_i\left(\frac{p_i^2}{q_i}-2p_i+q_i\right) \\
&=\sum_{i:p_i>0}\frac{p_i^2}{q_i}-1,
\end{aligned}
\label{eq:sparse}
\end{equation}
where the final equality uses $\sum_i p_i=\sum_i q_i=1$. Terms with $p_i=0$
disappear from the final sum. The query therefore evaluates the exact score
using only categories present in the current window.

Vocabulary aggregation can hide changes among raw values. For a group $G$,
define $P_G=\sum_{i\in G}p_i$ and $Q_G=\sum_{i\in G}q_i$.
Cauchy--Schwarz gives
\begin{equation}
\frac{(P_G-Q_G)^2}{Q_G}
\leq\sum_{i\in G}\frac{(p_i-q_i)^2}{q_i}.
\label{eq:coarsening}
\end{equation}
The contribution of a merged category cannot exceed the sum of the original
contributions. Increases and decreases among values inside \texttt{OTHER} can
cancel before scoring. RankShift remains exact for the deployed vocabulary,
while the finer raw-value representation can contain information lost through
aggregation. This result motivates hierarchical refinement of anomalous groups.

\vfill
\section{DuckDB Implementation}\label{app:query}

Listing~\ref{lst:duckdb} gives the complete DuckDB implementation evaluated in
the paper. Its parameters specify the smoothing mass, vocabulary size, score
precision, alert level, and number of returned drivers. The query constructs
reference shares, calculates exact sparse scores, assigns conformal p-values,
produces alerts, and ranks increasing category contributions.

\lstset{style=sql}
\begin{lstlisting}[
	caption={Complete DuckDB query used by the experiments.},
	label={lst:duckdb}
]
WITH counts AS MATERIALIZED (
  SELECT role,w,cat,COUNT(*)::DOUBLE cnt
  FROM events GROUP BY role,w,cat
), totals AS (
  SELECT role,w,SUM(cnt) n FROM counts
  GROUP BY role,w
), reference_counts AS (
  SELECT cat,SUM(cnt) r FROM counts
  WHERE role='ref' GROUP BY cat
), reference_total AS (
  SELECT SUM(r) r_total FROM reference_counts
), reference_q AS (
  SELECT cat,(r+$tau/$category_count)
    /((SELECT r_total FROM reference_total)+$tau) q
  FROM reference_counts
), observed AS MATERIALIZED (
  SELECT c.role,c.w,c.cat,c.cnt/t.n p,
    COALESCE(q.q,($tau/$category_count)
      /((SELECT r_total FROM reference_total)+$tau)) q
  FROM counts c JOIN totals t USING(role,w)
  LEFT JOIN reference_q q USING(cat)
  WHERE c.role IN ('cal','test')
), scored_raw AS MATERIALIZED (
  SELECT role,w,
    GREATEST(SUM(p*p/q)-1.0,0.0) score
  FROM observed GROUP BY role,w
), scored AS MATERIALIZED (
  SELECT role,w,score,
    ROUND(score,$score_key_decimals) score_key
  FROM scored_raw
), ranked AS (
  SELECT w,cat,(p-q)*(p-q)/q cell,
    ROW_NUMBER() OVER (PARTITION BY w
      ORDER BY (p-q)*(p-q)/q DESC,cat) driver_rank
  FROM observed
  WHERE role='test' AND p>q
), top_drivers AS (
  SELECT w,cat,cell,driver_rank FROM ranked
  WHERE driver_rank<=$top_k
), calibration AS (
  SELECT score_key calibration_score_key
  FROM scored WHERE role='cal'
), test_scores AS (
  SELECT w,score,
    (1.0+(SELECT COUNT(*) FROM calibration
      WHERE calibration_score_key>=scored.score_key))
    /((SELECT COUNT(*) FROM calibration)+1.0) p_value
  FROM scored WHERE role='test'
)
SELECT s.w,s.score,s.p_value,
  s.p_value<=$alpha is_alert,
  d.cat,d.cell,d.driver_rank
FROM test_scores s
LEFT JOIN top_drivers d USING(w)
ORDER BY s.w,d.driver_rank;
\end{lstlisting}

\newpage
\section*{Acknowledgments}
We thank Sudeep Agarwal, Mohit Suley, Greg Dunham, Mukul Sabharwal, and Euan
Grant (Microsoft) for leadership support of this work. We thank Zohair Shafi
(Northeastern University) for writing mentorship and guidance. We also thank
Abhishek Gupta, Varun Chawla, and the Azure Front Door, Web Application
Firewall, MAI/Bing Fundamentals, Bing Defense, and Bot Detection teams at
Microsoft for production-traffic access and operational feedback that informed
this work.

{\scriptsize
\paragraph{AI use disclosure.}
Portions of this manuscript and its accompanying artifacts (code, empirical
results, statistical tests, and figures) were generated with assistance from a
generative AI assistant. The author directed all experiments, reviewed all
outputs, and is responsible for the final content.

\paragraph{Disclaimer.}
Some of the information in this document relates to pre-released content which
may be subsequently modified. Microsoft makes no warranties, express or
implied, with respect to the information provided here. This document is
provided ``as-is''. Information and views expressed in this document, including
URL and other Internet Web site references, may change without notice.
\textcopyright\ 2026 Microsoft. All rights reserved.
\par}

\bibliographystyle{IEEEtran}
\bibliography{refs}

\end{document}